\documentclass{article} 
\usepackage[dvipsnames]{xcolor}

\usepackage{iclr2022_conference,times}

\usepackage{amsmath,amsfonts,bm}

\def\eqref#1{equation~\ref{#1}}

\def\1{\bm{1}}

\DeclareMathAlphabet{\mathsfit}{\encodingdefault}{\sfdefault}{m}{sl}
\SetMathAlphabet{\mathsfit}{bold}{\encodingdefault}{\sfdefault}{bx}{n}

\usepackage{hyperref}
\usepackage{url}
\usepackage[pdftex]{graphicx}

\title{BioLeaF: A Bio-plausible Learning Framework for Training of Spiking Neural Networks}

\author{Yukun Yang \thanks{Y. Yang is an incoming Ph.D. student to University of California, Santa Barbara.}, Peng Li  \\
Department of Electrical \& Computer Engineering\\
University of California, Santa Barbara\\
Santa Barbara, CA 93106, USA \\
\texttt{\{yukunyang,lip\}@ucsb.edu} \\

}

\iclrfinalcopy 
\begin{document}

\maketitle

\begin{abstract}
Our brain consists of biological neurons encoding information through accurate spike timing, yet both the architecture and learning rules of our brain remain largely unknown. Comparing to the recent development of backpropagation-based (BP-based) methods that are able to train spiking neural networks (SNNs) with high accuracy, biologically plausible methods are still in their infancy. In this work, we wish to answer the question of whether it is possible to attain comparable accuracy of SNNs trained by BP-based rules with bio-plausible mechanisms. We propose a new bio-plausible learning framework, consisting of two components: a new architecture, and its supporting learning rules. With two types of cells and four types of synaptic connections, the proposed local microcircuit architecture can compute and propagate error signals through local feedback connections and support training of multi-layers SNNs with a globally defined spiking error function. Under our microcircuit architecture, we employ the Spike-Timing-Dependent-Plasticity (STDP) rule operating in  local compartments  to update  synaptic weights and achieve supervised learning in a biologically plausible manner. Finally, We  interpret the proposed framework from an optimization point of view and show the equivalence between it and the BP-based rules under a special circumstance. Our experiments show that the proposed framework demonstrates learning accuracy comparable to BP-based rules and may provide new insights on how learning is orchestrated in biological systems.  

\end{abstract}
\section{Introduction}

Thanks to greater computing power, deep learning has gained remarkable achievements in recent years \citep{Hinton06, Bengio+chapter2007, schmidhuber2015deep, goodfellow2016deep}. However, learning by backpropagation (BP) \citep{rumelhart1986learning} is still the most popular method, which is generally believed impossible to be implemented in our brains \citep{illing2019biologically}. As compared to deep neural networks (DNNs), our brain, the only known true intelligence system, is more energy efficient \citep{von2012computer}, robust \citep{deneve2017brain, qiao2019defending}, and capable of achieving life-long learning \citep{parisi2019continual}, online learning \citep{lobo2020spiking}, logic reasoning \citep{monti2012logic}, and has many other advantages \citep{raichle2001default}. The development of artificial intelligence (AI) may benefit from investing in how our brain works.

Our brain is a complex system consisting of neurons that communicate with each other through spikes. Therefore, people tried to use simplified spiking neurons to form a network that mimic the function of our brain. Such spiking neural networks (SNNs) can naturally exploit spatio-temporal data with each neuron's internal temporal dynamics \citep{yang2021backpropagated}, and save orders of magnitude of less energy when running on neuromorphic hardwares \citep{davies2018loihi, kim2020spiking, davies2021advancing}. However, the training of SNNs is difficult.

Recent developments of the direct training methods of SNNs mainly diverge into two streams: BP-based rules and bio-plausible rules \citep{hao2020biologically}.

BP-based learning rules include: the activation-based surrogate gradient methods \citep{zenke2018superspike, shrestha2018slayer, wu2018spatio}, the timing-based methods \citep{zhang2020temporal}, the combination of both \citep{kim2020unifying}, and the recently proposed neighborhood aggregation method - NA \citep{yang2021backpropagated}. These BP-based methods gained great performance improvement and helped SNNs to be implemented on real-world problems, yet their biological plausibility remains unresolved: the co-existence of both forward and backward signals requires a neuron to fire two sets of uncorrelated signals from the same neuron body, which is not bio-plausible.


While the other branch, the bio-plausible learning rules, represented by the STDP \citep{taylor1973problem, levy1983temporal} and the Widrow-Hoff (WH) \citep{widrow1960adaptive} rules, adjusts parameters using local plasticity only. 

The STDP learning rule is built upon the Hebbian learning rule, which can be informally described as: "Cells that fire together, wire together" \citep{hebb1949organization}. Following this rule, STDP adjusts synaptic weights by evaluating the timing correlation: If a presynaptic neuron fires a few milliseconds before a postsynaptic neuron, meaning this presynaptic spike contributes to the firing of the postsynaptic neuron, their connection is strengthened (causal), or called long-term potentiation. Whereas the opposite temporal order results in long-term depression (acausal). Although STDP demonstrates its potential usefulness in both supervised and unsupervised manners, it is unlikely that STDP works alone: Strengthened connection makes the firing activity of a pair of neurons more synchronized, and vice versa. Due to the existence of the positive feedback loop, one needs to introduce additional tricks to stabilize the learning process - such as winner-takes-all (WTA) \citep{nessler2013bayesian,diehl2015unsupervised,kheradpisheh2018stdp,saunders2018stdp}, weights normalization \citep{ferre2018unsupervised}, weights clamping \citep{diehl2015unsupervised, kheradpisheh2018stdp, lee2018training, saunders2018stdp}, layer-by-layer training \citep{kheradpisheh2018stdp}, and others \citep{panda2016unsupervised}. 

In comparison, the WH-based learning algorithms, represented by ReSuMe\citep{ponulak2010supervised} and SPAN \citep{mohemmed2012span}, are able to train a spiking neuron to generate spikes with accurate timing, and do not need additional tricks as STDP does. The WH learning rule \citep{widrow1960adaptive} is a special case of the gradient descent rule where the least mean square loss is applied. \cite{ponulak2010supervised} presented a spiking analogy to the classical WH rule for spiking neuron models, and their rule can be interpreted as an STDP-like process between a presynaptic spike train and a postsynaptic error signal. However, previous WH-based methods are constrained to train a single layer SNN since it has difficulty in propagating the teaching signal to previous layers. \cite{sporea2013supervised} extend the single layer WH-based learning rule - ReSuMe \citep{ponulak2010supervised} onto multi-layers networks through BP-liked error propagation, which is practical but deviates from the original intention of exploring bio-plausible mechanisms.

\begin{figure}[t]
    \centering
    \includegraphics[width=0.95\linewidth]{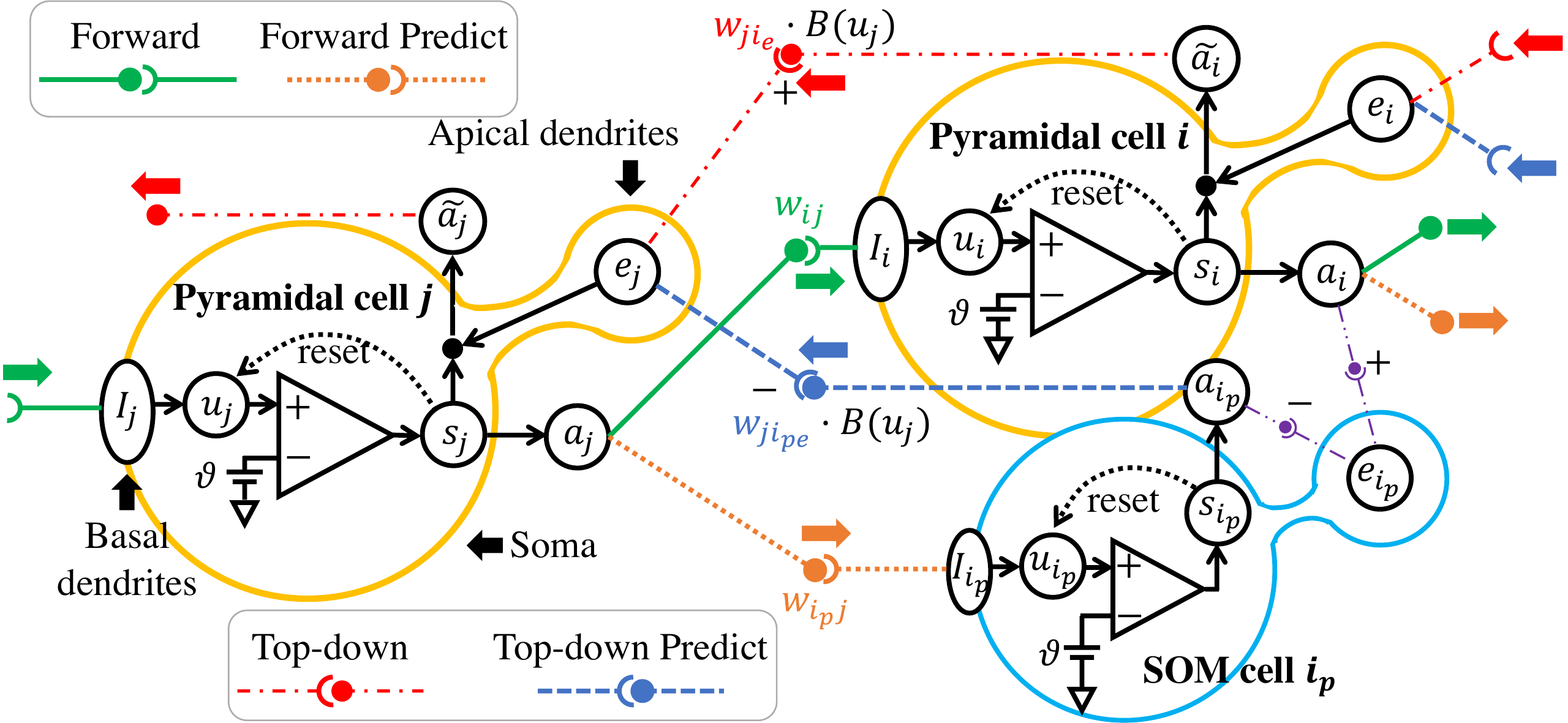}
    \caption{Our proposed Microcircuit architecture.}
    \label{fig:framework}
    \vspace{-0.5cm}
\end{figure}

In this work, we propose a \textbf{bio}-plausible \textbf{lea}rning \textbf{f}ramework - BioLeaF, underpinned by two key components: \textbf{1)} a microcircuit architecture consisting of two types of spiking neurons and four types of synapses as shown in Figure \ref{fig:framework}, and \textbf{2)} the STDP-based learning rules built upon our architecture.

The architecture is inspired from the predict-coding-based algorithms \citep{rao1999predictive, stefanics2014visual}. Previous works proposed several predictive-coding-inspired microcircuit architectures to realize BP-liked learning on rate-based neurons \citep{bastos2012canonical, whittington2017approximation, sacramento2018dendritic}, where all neurons communicate with each others through continuous currents, and no explicit temporal point processes or spiking behaviors are included. This simplified setting limits the discussion of the widely used  bio-plausible learning rules defined by the spike-timing correlation like STDP rule and WH rule. Our architecture differs from them in both the neuron and the synapse models. We include the more bio-plausible spiking leaky integrate-and-fire (LIF) neuron \citep{gerstner2002spiking} and synapses models that transmit discontinuous spikes into our architecture.

The architecture consists of two types of spiking cells - pyramidal cells and somatostatin-expressing (SOM) cells \citep{petreanu2009subcellular, larkum2013cellular}. Each pyramidal cell $i$ has a paired SOM cell $i_p$ to predict its firing activity one-on-one through the same current inputs $a_j$, $j=1\cdots N$, where $N$ is the total number of presynaptic neurons. The prediction mismatch, also interpreted as a surprise or free energy \citep{friston2010free}, is transmitted through top-down connections to all presynaptic neurons' apical dendrites, and acts as their error signals. A pyramidal cell's top-down output $\tilde{a}$ is modulated by its error signals, whereas a SOM cell's top-down output $a_p$ is not. Therefore, without knowing the signal in $i$'s apical dendrite, SOM cell $i_p$'s prediction can only cancel out part of $i$'s output, which leaves the error-related signal on $j$'s apical dendrites. The summation of all top-down signals will cancel out each others pair by pair, and leaves the total error-related signals onto $j$, from where the layer-by-layer error backpropagation is realized.


The learning rule built upon our microcircuit architecture is the standard STDP rule as defined in \cite{levy1983temporal} with specialized choices of pre/postsynaptic signals. As comparing to typical SNNs' architectures \citep{shrestha2018slayer, wu2018spatio, yang2021backpropagated}, which only have forward connected weights, we introduce three additional types of weights: forward predict, top-down, and top-down predict as in Figure \ref{fig:framework} to support the bio-plausible learning. Therefore, to update different types of synaptic weights, our STDP updating rules need to be defined between pairs of presynaptic and postsynaptic signals locates in different components of our microcircuit architecture. A presynaptic signal is an output spike train located in the presynaptic neuron like $s_j$ for both weights $\textcolor{ForestGreen}{w_{ij}}$ and $\textcolor{Orange}{w_{{i_p}j}}$, or $s_{i_p}$ for $\textcolor{NavyBlue}{w_{j{i_{pe}}}}$ following \cite{levy1983temporal}, and a postsynaptic signal is an error signal located in the postsynaptic neuron like $e_i$ for weights $\textcolor{ForestGreen}{w_{ij}}$ and $e_j$ for $\textcolor{NavyBlue}{w_{j{i_{pe}}}}$.

More generally, we analytically show that the the proposed framework is equivalent to the BP-based learning rules under certain settings. To derive a BP-based learning rules which propagates continuous-valued loss signal through discontinuous all-or-none firing activity, some approximation methods are applied following previous works \citep{shrestha2018slayer, wu2018spatio, yang2021backpropagated}. Yet such approximation surprisingly aligned with the standard STDP rules under our microcircuit architecture with only minor differences. We empirically build a 2-layers toy example to evaluate the learning ability of BioLeaF. Deeper than a single layer breaks the limit of the previous WH-based learning rules. Then, by benchmarking on the datasets including MNIST \citep{lecun1998mnist} and CIFAR10 \citep{krizhevsky2009learning}, the proposed BioLeaF also exhibits comparable accuracy with other BP-based methods when extended to multi-layers deep SNNs.

\section{Microcircuit Architecture}
\subsection{Spiking Neuron Model}
Both pyramidal cells and SOM cells are modeled by the leaky integrate-and-fire (LIF) neuron model, which is one of the most prevalent choices for describing dynamics of spiking neurons. 

\subsubsection{LIF Neuron Model}
The dynamics of the neuronal membrane potential $u$ of neuron $i$ in layer $l$ is described by:
\begin{equation}
    \tau_m\frac{d {u}_i^{(l)}(t)}{d t}  = -{u}_i^{(l)}(t)+ I_i^{(l)}(t)
    + \eta_i^{(l)}(t),
    \label{eq:pyramidal}
\end{equation}
where $I_i^{(l)}$ is the total input of synaptic currents, and $\eta_i^{(l)}(t)$ denotes the reset function. A spiking neuron reset its membrane potential from threshold $\vartheta$ to the resting potential $v_{\rm rest}$ (we set $v_{\rm rest}=0$) each time when it fires a spike. We model $\eta_i^{(l)}(t)$ as the time convolution (*) between a reset kernel $\nu$ and the neuron's output spike train ${s}_i^{(l)}$: $\eta_i^{(l)}(t) = (\nu \ast {s}_i^{(l)}) (t)$. The reset kernel $\nu(t)=-\vartheta\delta(t)$. The amount of resetting is equal to the threshold $\vartheta$ (we set $\vartheta=1$), and $\delta(t)$ is the Dirac delta function. The neuron's output spike train is also modeled by a serious of delta functions as:${s}_i^{(l)}=\sum_f\delta(t-t_{i(f)}^{(l)})$. Here, $t_{i(f)}^{(l)}$ represent the firing time of the $f^{th}$ spike of neuron $i$ in layer $(l)$. An output spike is generated once the membrane potential reaches the threshold $\vartheta$. Following \cite{shrestha2018slayer}, we define the spike function as:
\begin{equation}
f_s(u):u\rightarrow s, s(t):=s(t)+\delta(t-t_{(f+1)}),
    t_{(f+1)}={\rm min}\left\{ t: u(t)=\vartheta, t>t_{(f)}\right \}.
\end{equation}

\subsection{Synaptic Currents}
\label{sec:synapse}

We model the general total input current on neuron i as: $I_i(t) = \sum_{j}{w_{ij}}\sum_f\alpha_{ij}(t-t_{j(f)})$.
The modeled total input is the weighted sum over all current pulses:
\begin{equation}
    \alpha_{ij}\left(t-t_{j(f)}\right) = g_{ij}\left(t-t_{j(f)}\right)\cdot \left[E_{\rm syn}-u_i(t)\right],
\end{equation}
where $E_{\rm syn}$ is the reversal potential for the synaptic current. 
We set $E_{\rm syn}\gg\vartheta$ in all types of synapses \citep{destexhe1998kinetic}, so the membrane potential dependency can be neglected, and the term $[E_{\rm syn}-u_i(t)]$ can be treated as a constant and absorbed into weights. 
$g_{ij}(t-t_{j(f)})$ is the synaptic conductance change. We modeled it following \cite{eyal2018human}, but simplified the double exponential function to a single exponential decaying function and have:
\begin{equation}
    g_{ij}\left(t-t_{j(f)}\right)=B_i(t)\cdot ({1}/{\tau_s})\cdot exp\left[-\left(t-t_{j(f)}\right)/{\tau_s}\right],
    \label{eq:g}
\end{equation}
where $B_i(t)$ is a membrane potential dependent gating function.  The synapses in different location have different properties, where the two types of synapse we apply are named as the \emph{forward-related-type (F-type)} and the \emph{error-related-type (E-type)}. We introduce them one by one as following:

\textbf{F-Type Synapses}:

The connections from the outputs of pyramidal cells in one layer to the basal dendrites of pyramidal cells in their next layer carrying important feature information build up the main architecture in a spiking neural networks. When training is finished, only these forward connections are needed to realize inference. We implement these forward connections $\textcolor{ForestGreen}{w_{ij}^{(l)}}$ together with their paired forward predict connections $\textcolor{Orange}{w_{i_pj}^{(l)}}$ with the F-type synapses.

$B_i(t)$ is set to 1 (voltage independent conductance) as a general setting for AMPA-based conductance \citep{eyal2018human}. We follow this setting and the input current is then simplified to a widely-used alpha function with time constant $\tau_s$:
\begin{equation}
    \alpha_{ij}\left(t-t_{j(f)}\right) = ({1}/{\tau_s})\cdot exp\left[-\left(t-t_{j(f)}\right)/{\tau_s}\right].
    \label{eq:Ftype1}
\end{equation}
Under which, the synaptic current is independent from neuron $i$, and all postsynaptic current (PSC) generated from neuron $j$ can be uniformly expressed by one variable $a_j$ as:
\begin{equation}
    a_j(t)=\sum_f\alpha_{ij}(t-t_{j(f)})=(s_j*\epsilon)(t),~~~~\epsilon(t) = ({1}/{\tau_s})\cdot exp{(-t/{\tau_s})} H(t),
    \label{eq:Ftype2}
\end{equation}
where (*) represents the time convolution. $\epsilon(\cdot)$ is the impulse response. $H(\cdot)$ represents the Heaviside step function: $H(t)=1, t\geq 0$ and  $H(t)=0, t<0$.

A fully connected layer can be described through the current flows as:
\begin{equation}
    I_i^{(l)}(t) = \sum_{j=1}^{N^{(l-1)}}\textcolor{ForestGreen}{w_{ij}^{(l)}}\sum_f\alpha_{ij}(t-t_{j(f)}^{(l-1)}) = \sum_{j=1}^{N^{(l-1)}}\textcolor{ForestGreen}{w_{ij}^{(l)}}a_j^{(l-1)}(t).
\end{equation}
Other layers like convolution layers can be easily converted to fully connected layers. Similarly, the total current of SOM cell $I_{i_p}^{(l)}$ is modeled as: $I_{i_p}^{(l)}(t) =\sum_{j=1}^{N^{(l-1)}}\textcolor{Orange}{w_{i_pj}^{(l)}}a_j^{(l-1)}$, 
where the footnote $p$ represent the predict-related, or the SOM-related variables.

\textbf{E-Type Synapses}:

All other connections are modeled by E-type synapses, which differs from the F-type synapses by their postsynaptic voltage dependent property. We model the voltage dependent gating function $B_i(t)$ like how prior works model the NMDA-based synapses \citep{eyal2018human}. $B_i(t)$ has a shape that peaks when the postsynaptic cell's membrane potential $u_i(t)$ reaches the threshold $\vartheta$, and decrease as $u_i(t)$ moves away from $\vartheta$. Such a shape acts as surrogate derivative function when compared to BP-based methods, which will be fully discussed in the following section.

We define $B_i(t)$ of our synapses on the apical dendrites as following:
\begin{equation}
    B_i(t) = \frac{g_{\rm max}}{1+exp\left[-k \left(u_i(t)-u_0\right)\right]\cdot[{\rm Mg^{2+}}]\cdot n},
\end{equation}
where the extracellular magnesium concentration $[{\rm Mg^{2+}}]$ was 1 mM in the model. We shift the voltage dependency of $B_i(t)$ by $u_0=\vartheta$, and tune the parameters $g_{\rm max}$, $k$ and $n$ to fits the function $B$ into our simplified setting where $v_{\rm rest}=0$, and $\vartheta=1$.

The error backpropagation is achieved by the corporation between pairs of pyramidal cells and SOM cells. Higher level pyramidal cells' top-down output currents are coupled by the error signals located in their apical dendrites. We model this coupling effect by the current sum of both a neuron's PSC and its error signal as: $\widetilde{a}_i^{(l)}(t) = a_i^{(l)}(t) + e_i^{(l)}(t)$. A pyramidal cell's top-down connection contributes positively with weights $\textcolor{Red}{w_{ji_e}^{(l)}}$ onto previous layers' pyramidal cells' apical dendrites, and its paired SOM cell contribute negatively with weights $\textcolor{NavyBlue}{w_{ji_{pe}}^{(l)}}$, where the footnotes $p$ stands for SOM-related, and the footnotes $e$ stands for error-related. We express the total error signals on the pyramidal cell $j$'s apical dendrites as:
\begin{equation}
    e_j^{(l-1)}(t)=B_{j}^{(l-1)}(t)\left[\sum_{i=1}^{N^{(l)}}\left(\textcolor{Red}{w_{ji_e}^{(l)}}\widetilde{a}_i^{(l)}(t)-\textcolor{NavyBlue}{w_{ji_{pe}}^{(l)}}a_{i_p}^{(l)}(t)\right)\right],
    \label{eq:pyramidal_error}
\end{equation}
For the output layer, the apical dendrites receives the one-on-one error signal from higher brain areas to realize supervised learning.
\begin{equation}
    e_i^{(l_N)}(t) = B_{i}^{(l_N)}(t)\left[a_i^{\rm target}(t) - a_i^{(l_N)}(t)\right]
    \label{eq:bio-error}
\end{equation}
In our framework, SOM cells mimic the behavior of the same layer's pyramidal cells one-on-one. Therefore, a one-to-one nudging signal from a pyramidal cell to its corresponding SOM cell (as the dashed purple connections in Figure \ref{fig:framework}) are needed. Together with the negative feedback output currents that SOM cells generated themselves, we get the local error signals $e_{i_p}^{(l)}$ of an SOM cell $i$ in layer $(l)$:
\begin{equation}
    e_{i_p}^{(l)}(t) = \alpha_{i_pj}^{(l)}(t)-\alpha_{i_pi_p}^{(l)}(t) =B_{i_p}^{(l)}(t) \left[a_i^{(l)}(t)-a_{i_p}^{(l)}(t)\right].
    \label{eq:SOM_error}
\end{equation}
Although this simplified one-to-one setting impose special constrains on the neural network's connectivity, the recent monosynaptic experiments confirm that the SOM cells do receive top-down connections which may encode such teaching information \citep{leinweber2017sensorimotor}. 

The SOM cells differs from the pyramidal cells by the sign of their output currents, but it does not mean that we fix the type of a cell to be excitatory or inhibitory. Instead, we allow the synapses' connection weights to move across zero freely and change the sign of its current, which is a general setting in previous works \citep{ponulak2010supervised, sacramento2018dendritic}. When currents from both SOM cells and pyramidal cells summed together into the SOM cells or the apical dendrites as in (\ref{eq:SOM_error}) and (\ref{eq:pyramidal_error}), a minus sign is added for SOM cell's outputs.
\section{Bio-plausible Learning Rules}

\subsection{Spike-Timing-Dependent-Plasticity (STDP)}
We first introduce a popular version of the STDP rules: $\Delta w_{ij}=\sum_m\sum_n \kappa_{\rm STDP}(t_{i}^{m}-t_{j}^{n})$, where $\kappa_{\rm STDP}$ is the STDP kernel function, which is modeled by the two-sides exponential decaying function defined as:
\begin{equation}
    \kappa_{\rm STDP}(\Delta t) = \left\{
    \begin{array}{lr}
    A^+ \cdot exp(-{\Delta t}/{\tau_+}),& \Delta t>0 \\
     -A^- \cdot exp({\Delta t}/{\tau_-}),& \Delta t<0
    \end{array}\right.
    \label{eq:STDPkernel}
\end{equation}
Meantime, we define a reversed STDP kernel function $\tilde{\kappa}_{\rm STDP}(t) = \kappa_{\rm STDP}(-t)$. Recalling that all spike trains are a serious of delta function ${s}=\sum_f\delta(t-t_{(f)})$, and considering the delta function's sampling property $\int [f(t) \delta(t-T)]~dt=f(T)$, we rewrite the STDP updating rule into two equivalent forms:
\begin{equation}
    \Delta w_{ij}=\eta_{ij}\int (s_j \ast \kappa_{\rm STDP})(t) \cdot s_i(t)~dt
                =\eta_{ij}\int s_j(t) \cdot (s_i\ast \tilde{\kappa}_{\rm STDP})(t)~dt,
\label{eq:int_STDP}
\end{equation}
where $\eta_{ij}$ is the learning rate. Although $\eta_{ij}$ can be absorbed into $A^+$ and $A^-$, we explicitly define it for clarity.

Since the WH rule can be interpreted as an STDP-like process between the presynaptic spike trains $s_j$ and an error signal $e_i:=s_i^{\rm teach}-s_i$. In this work, we represent both the STDP rule and the WH rule uniformly as ${\rm STDP}(\cdot,\cdot)$, where the STDP rule is: ${\rm STDP}(s_{pre},s_{post})$, and the WH rule is: ${\rm STDP}(s_{pre},e_{post})$. In addition,  WH rule's kernel function $\kappa_{\rm WH}$ is usually equal to $\kappa_{\rm STDP}$ as in (\ref{eq:STDPkernel}). Defining the reverse kernel function $\tilde{\kappa}_{\rm WH}(t) = \kappa_{\rm WH}(-t)$, we have the weight updating rule of the WH rule as:
\begin{equation}
    \Delta w_{ij}=\eta_{ij}\int (s_j \ast \kappa_{\rm WH})(t) \cdot e_i(t)~dt
                =\eta_{ij}\int s_j(t) \cdot (e_i \ast \tilde{\kappa}_{\rm WH})(t)~dt.
\label{eq:int_WH}
\end{equation}
It is noteworthy that the WH-based rules provides a fixed point in the weight space, $\Delta w_{ij}=0$ when $e_i=0$, which means $s_i^{\rm teach}=s_i$. It is a global positive attractor under certain conditions \citep{ponulak2010supervised}.

\subsection{Our bio-plausible learning rules}
We first conclude our synaptic learning rules as following:
\begin{equation}
    \Delta \textcolor{ForestGreen}{w_{ij}^{(l)}} = {\rm STDP}(s_j^{(l-1)}, e_i^{(l)})
    =\eta_{ij}\int\left( s_j^{(l-1)}(t)\ast  \kappa_{\rm STDP}\right) (t)\cdot e_i^{(l)}(t) dt
\label{eq:forward_weights}
\end{equation}
\begin{equation}
    \Delta \textcolor{Orange}{w_{i_pj}^{(l)}} = {\rm STDP}(s_j^{(l-1)}, e_{i_p}^{(l)})
    =\eta_{i_pj}\int \left(s_j^{(l-1)}(t)\ast  \kappa_{\rm STDP}\right)(t)\cdot e_{i_p}^{(l)}(t) dt
\label{eq:predict_weights}
\end{equation}
\begin{equation}
    \Delta \textcolor{NavyBlue}{w_{ji_{pe}}^{(l)}} = {\rm STDP}(s_{i_p}^{(l)}, e_j^{(l-1)})
    =\eta_{ji_{pe}}\int \left(s_{i_p}^{(l)}(t)\ast \kappa_{\rm STDP}\right)(t)\cdot e_j^{(l-1)}(t)dt
\label{eq:error_weights}
\end{equation}
where $\eta_{ij},\eta_{i_pj}$, and $\eta_{ji_{pe}}$ are three different learning rates for these three types of synaptic connections. In the above three equations, all the adjusting rules of a synapse can be described as a STDP process between a presynaptic current and a postsynaptic error signal, which exploits great biological plausibility. The top-down synaptic weights $\textcolor{Red}{w_{ji_{e}}^{(l)}}$ are set to be equal to the bottom-up forward weights $\textcolor{ForestGreen}{w_{ij}^{(l)}}$ for simplicity. One can also try to fixed the top-down weights $\textcolor{Red}{w_{ji_{e}}^{(l)}}$ as randomly initialized values following the idea of the feedback alignment \citep{sacramento2018dendritic}.

We interpret each type of synapses' learning rule as following:

In (\ref{eq:forward_weights}), $\textcolor{ForestGreen}{w_{ij}^{(l)}}$ represents the forward weights. It's fixed point $e_i^{(l)}(t)=B_{i}^{(l_N)}(t)[a_i^{\rm target}(t) - a_i^{(l_N)}(t)]=0$ forms a positive attractor to minimize the error signal. When training is converged, we expect $a_i^{(l_N)}\approx a_i^{\rm target}$ for all pyramidal cells $(i = 1,\dots, N^{(l)}), (l=1,\dots,l_N)$.

In (\ref{eq:predict_weights}), $\textcolor{Orange}{w_{i_pj}^{(l)}}$ stands for the predictive connections. Its fixed point in our learning rule is $e_{i_p}^{(l)}(t) = B_{i_p}^{(l)}(t) [a_i^{(l)}(t)-a_{i_p}^{(l)}(t)] = 0$, which means $a_i^{(l)}=a_{i_p}^{(l)}$, and $s_i^{(l)}=s_{i_p}^{(l)}$. As compared to (\ref{eq:forward_weights}), the learning rule of (\ref{eq:predict_weights}) does not have a fixed target signal, but needs to follow the continuously changing behavior of each SOM cell's corresponding pyramidal cell during training. 

The learning rule (\ref{eq:error_weights}) minimizes the norm of $e_j^{(l-1)}(t)$ in (\ref{eq:pyramidal_error}) through adjusting $\textcolor{NavyBlue}{w_{ji_{pe}}^{(l)}}$, which is equivalent to solving a linear equation: $\sum_{i=1}^{N^{(l)}}\textcolor{NavyBlue}{w_{ji_{pe}}^{(l)}}a_{i_p}^{(l)}(t)=\sum_{i=1}^{N^{(l)}}\textcolor{Red}{w_{ji_e}^{(l)}}(a_i^{(l)}(t)+e_i^{(l)}(t))$.

Considering the Current-Time as a continuous $f(X)\leftrightarrow X$ function space, where all SOM cells' PSCs $a_{i_p}^{(l)}, (i=1,\dots N_l)$ form a basis of the space. Then the goal here is to restore the summed currents of all pyramidal cells in layer $(l)$ through these basis functions. When a predictive weight $\textcolor{Orange}{w_{i_pj}^{(l)}}$ is well adjusted, $a_i^{(l)}$ approximately equals to $a_{i_p}^{(l)}$, then an obvious solution to the equation above will be letting $\textcolor{NavyBlue}{w_{ji_{pe}}^{(l)}}$ equal to $\textcolor{Red}{w_{ji_e}^{(l)}}$, if we consider the error currents $e_i^{(l)}(t)$ as the orthogonal uncorrelated signal to the basis.
\section{Relationship to the BP Algorithm}

\subsection{Backpropagation Flow}
Since the SOM cells are auxiliary in our architecture, we only introduce how backpropagation works in a general SNN without SOM cells.

Consider a general loss function $L=\int_t E(t) dt$, which is defined on the output layers' PSC $a_i^{(l_N)}(t), ~i=1,\dots,N^{(l)}$, where $l_N$ is the total number of layers. The differentiable property of a loss function requires that $\partial L/\partial a^{(l_N)}(t)$ exist. We name the partial derivative on the layer $(l)$'s PSC as:$ d_i^{(l)}(t) := \frac{\partial L}{\partial  a_i^{(l)}(t)}.$

Taking the $l2$ distance between two PSCs, or the van Rossum distance \citep{van2001novel} between two spike trains, as an example:
\begin{equation}
    L=\frac{1}{2}\int \left[\left(a_i^{\rm target}-a_i^{(l_N)}\right)(t)\right]^2dt=\frac{1}{2}\int \left[\left((s_i^{\rm target}\ast\epsilon)-(s_i^{(l_N)}\ast\epsilon)\right)(t)\right]^2dt
\end{equation}

When computing the gradient of loss with respect to the synaptic weight of the last layer $w^{(l_N)}_{ij}$:
${\partial L}/{\partial w^{(l_N)}_{ij}} = \int d_i^{(l_N)}(t)\left(\epsilon \ast {\partial s_i^{(l_N)}}/{\partial w^{(l_N)}_{ij}} \right)(t) dt$, one may find the derivative ${\partial s_i^{(l_N)}}/{\partial w^{(l_N)}_{ij}}$ is ill-defined due to spiking neurons' discontinuous all-or-none firing activities. Following \citep{zenke2018superspike}, we substitute this term by: $(\sigma'(u_i^{(l_N)})\cdot{\partial u_i^{(l_N)}}/{\partial w^{(l_N)}_{ij}})$, and further approximate the term $(\partial u_i^{(l_N)}/\partial w^{(l_N)}_{ij})\approx a_j^{(l_N-1)}$ by omitting the temporal dependency of membrane potentials. We have the weights updating rule as:
\begin{multline}
    \Delta w^{(l_N)}_{ij} = -\eta \frac{\partial L}{\partial w^{(l_N)}_{ij}} = \eta \int -d_i^{(l_N)}(t)\left(\epsilon \ast (\sigma'(u_i^{(l_N)})\cdot a_j^{(l_N-1)})\right)(t) dt \\
    \approx \eta \int \underbrace{\left(-d_i^{(l_N)}\sigma'(u_i^{(l_N)})\right)(t)}_{\rm post} \cdot \underbrace{\left(\epsilon \ast (s_j^{(l_N-1)} \ast \epsilon))\right)(t)}_{\rm pre}
    = \eta \int e_i^{(l_N)}(t) \cdot \left(s_j^{(l_N-1)} \ast \kappa_{\rm BP}\right)(t),
\label{eq:BP}
\end{multline}
where $\eta$ is the learning rate, $e_i^{(l_N)}:= -d_i^{(l_N)}\sigma'(u_i^{(l_N)})$, and the kernel function of BP algorithm $\kappa_{\rm BP}(t) := (\epsilon\ast\epsilon)(t)=\left[(t/\tau_s^2)\cdot exp(-t/\tau_s)\cdot H(t)\right]$. In the second line, we approximately switch the order of time convolution and product to separate variables of presynaptic neurons and postsynaptic neurons. Then to further propagate the gradient to previous layers, we calculate the partial derivative of loss with respect to hidden layer's PSCs as:
\begin{equation}
    d_j^{(l-1)}(t) = \sum_{i=1}^{N^{(l)}} \left(d_i^{(l)}\ast \epsilon\right)(t) \cdot \sigma'\left(u_i^{(l)}(t)\right) w_{ij}^{(l)} 
    \approx \sum_{i=1}^{N^{(l)}} w_{ij}^{(l)} \left(d_i^{(l)}\cdot \sigma'(u_i^{(l)})\right)(t) =-\sum_{i=1}^{N^{(l)}} w_{ij}^{(l)} e_i^{(l)},
\label{eq:BP_hidden}
\end{equation}
Here we omit the temporal dependency $(\ast \epsilon)$ for simplicity. Since BP is not our contribution, we discuss different omitting methods of all previous works and the detailed derivatives including the full dependency in the appendix. With $d_j^{(l-1)}(t)$ calculated, an hidden layer follows the same rule to propagate from $d_j^{(l-1)}$ to $({\partial L}/{\partial w^{(l-1)}_{jm}})$ as the output layer. 

\subsection{Comparing BP with our framework}
We analyses our bio-plausible learning rule from the optimization point of view. 

\begin{figure}[t]
    \centering
    \includegraphics[width=0.999\linewidth]{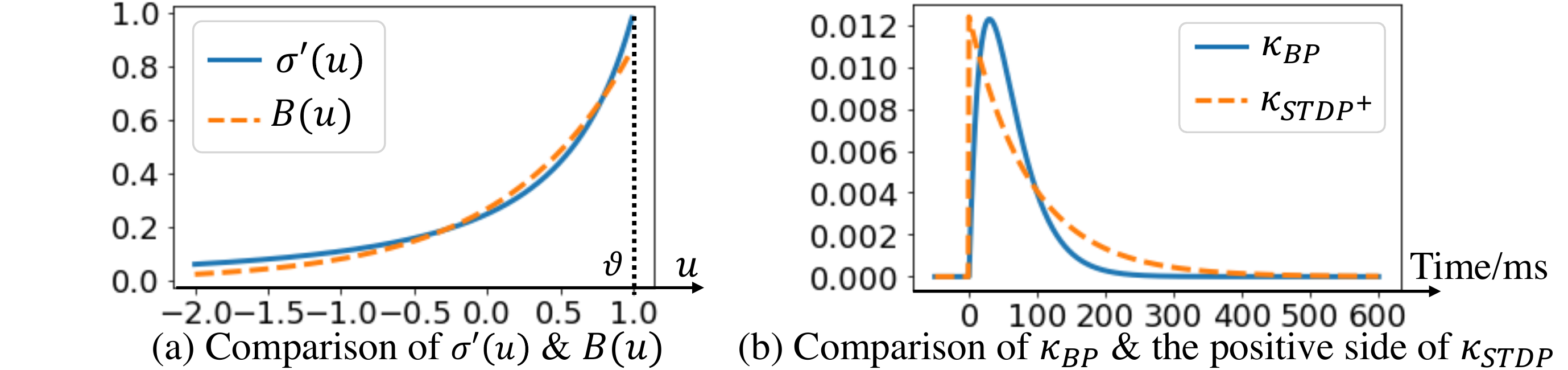}
    \caption{(a) $\sigma'(u) = 1/(1+|u-1|)^2)$ is defined following \cite{zenke2018superspike}, and the parameters of $B(u)$ used here are: $u_0=\vartheta=1$, $g_{\rm max}=109.45$, $k=1.18$ and $n=124.33$. (b) $\kappa_{\rm BP}=\epsilon \ast \epsilon$, with $\tau_s=30$ ms. The positive side of $\kappa_{\rm STDP}=A^+exp(-t/\tau_+)H(t)$ has parameters: $A^+=0.0124$, and $\tau_+=89.73$ ms.}
    \label{fig:similarity}
    \vspace{-0.5cm}
\end{figure}

In the output layer, the error signal in (\ref{eq:bio-error}) corresponds to $\left(-d_i^{(l_N)}\sigma'(u_i^{(l_N)})\right)$, which is the postsynaptic part in (\ref{eq:BP}). The term $-d_i^{(l_N)}$ corresponds to $[a_i^{\rm target}(t) - a_i^{(l_N)}(t)]$ in (\ref{eq:bio-error}), which implies that the equivalent loss function we apply for the bio-plausible rule is also the van Rossum distance. For other loss functions, one can safely substitute $[a_i^{\rm target}(t) - a_i^{(l_N)}(t)]$ with their own $(-d_i^{(l_N)})$. The auxiliary $\sigma'(u_i^{(l_N)})$ function corresponds to the voltage dependent gating function $B_{i}^{(l_N)}$. One may find that with proper parametrization they can almost overlap as shown in Figure \ref{fig:similarity} (a). Both of them reshapes the error signal depending on the neuron's membrane potential, where the gradients farther from the threshold are weakened.

With error signals clearly defined, we find both the BP-based learning rule (\ref{eq:BP}) and the previously described bio-plausible STDP-based learning rule (\ref{eq:forward_weights}) follow a surprisingly similar form: 
\begin{equation}
    \Delta w_{ij} \propto \int (s_{\rm pre}\ast \kappa)\cdot e_{\rm post}.
\end{equation}
As shown in Figure \ref{fig:similarity} (b), the shape of BP's kernel function $\kappa_{\rm BP}$ is highly similar to the positive side of the STDP kernel $\kappa_{\rm STDP}$, where the value peaks near zero, and decays gradually. Such equivalence gives theoretical analysis of what should the kernel looks like from the optimization point of view and provides possibly explanations of why the negative side of STDP learning rule usually dampen the performance, and usually been omitted in previous works to boost the performance \citep{ponulak2010supervised}. In the experiment part, we also ignored the negative side of STDP kernel.

When further propagating the gradient to previous layers, the more complex predictive-coding-inspired architectures are involved. As in \cite{sacramento2018dendritic}, we also name the ideal state where $\textcolor{ForestGreen}{w_{ij}^{(l)}}=\textcolor{Orange}{w_{i_pj}^{(l)}}=\textcolor{NavyBlue}{w_{ji_{pe}}^{(l)}}=\textcolor{Red}{w_{ji_e}^{(l)}}$ as \emph{self-predicting} (\emph{self-predicting} is needed theoretically, but not experimentally as shown in the next section). Under which, the summed error signal on the apical dendrites of a pyramidal cell in (\ref{eq:pyramidal_error}) are simplified to:
\begin{equation}
    e_j^{(l-1)}(t)=B_{j}^{(l-1)}(t)\sum_{i=1}^{N^{(l)}}\left(\textcolor{ForestGreen}{w_{ij}^{(l)}}(a_i^{(l)}(t)+e_i^{(l)}(t))-\textcolor{ForestGreen}{w_{ij}^{(l)}}a_{i}^{(l)}(t)\right)=B_{j}^{(l-1)}(t) \sum_{i=1}^{N^{(l)}}\textcolor{ForestGreen}{w_{ij}^{(l)}}e_i^{(l)}(t).
\end{equation}
Comparing to (\ref{eq:BP_hidden}), which propagates the negative weighted sum of $e_i^{(l)}$ to $d_j^{(l-1)}$, and further calculate $e_j^{(l-1)}=-d_j^{(l-1)}\sigma'(u_j^{(l)})$, our rule yields more symmetry, where the error signals $e$ flow through layers without any intermediate variables. These two rules are equivalent when pairing $B(u)$ to $\sigma'(u)$ as in Figure \ref{fig:similarity} (a).

We conclude their similarity and difference as following:
1) Both of the learning rules share a same form: $\Delta w_{ij} \propto \int (s_{\rm pre}\ast \kappa)\cdot e_{\rm post}$.
2) The surrogate derivatives  $\sigma'(u_i^{(l)})$ in BP is correspondingly achieved by $B_i^{(l)}$ of the E-type synapses in our framework.
3) The shape of BP's kernel function $\kappa_{\rm BP}$ is similar to the positive side of $\kappa_{\rm STDP}$.
4) For both rules, the error backpropagation between two layers are equivalent when our network in its \emph{self-predicting} state. 
And the one difference these two methods have is: The kernel function $\kappa_{\rm BP}$ only has the positive side $(t>0)$, but $\kappa_{\rm STDP}$ has double sides.

\section{Experimental Results}
\subsection{Universal Spike Train Approximator}
\begin{figure}[t]
    \centering
    \includegraphics[width=0.95\linewidth]{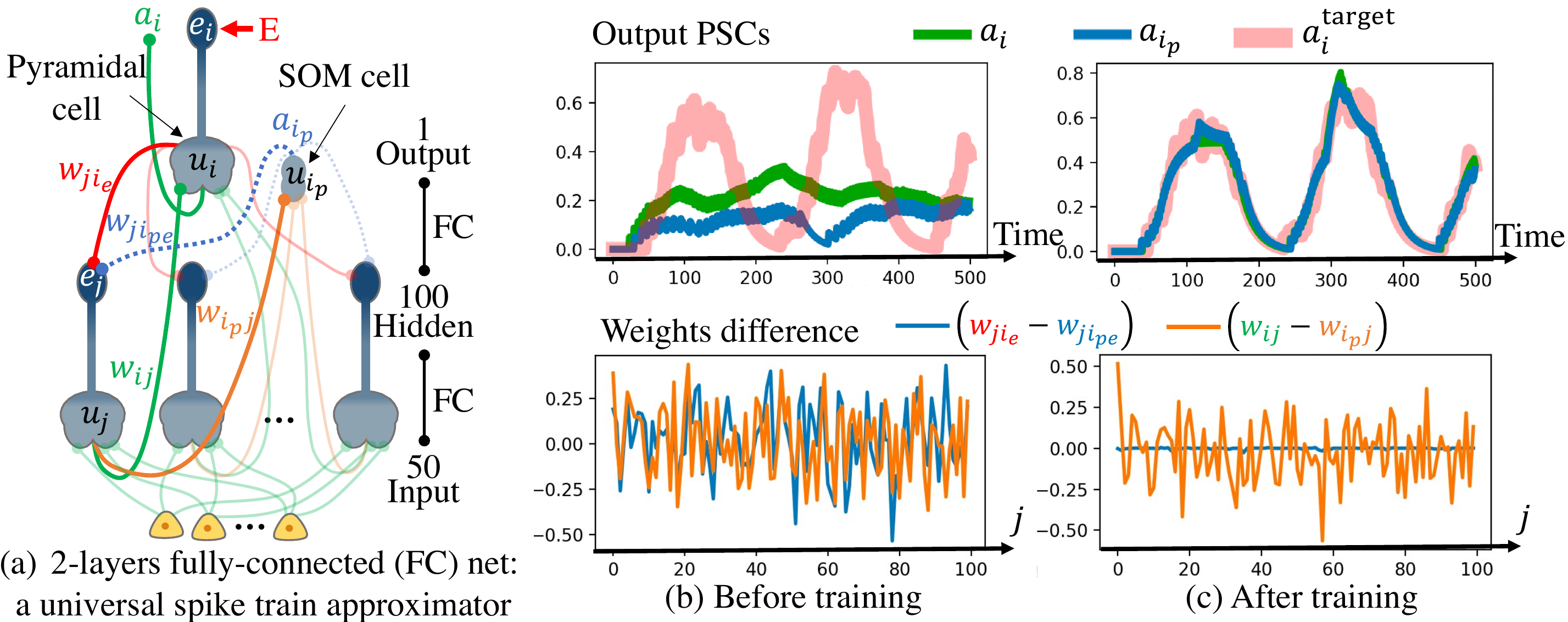}
    \caption{Universal spike train approximator experiment}
    \label{fig:toy_example}
    \vspace{-0.5cm}
\end{figure}

To test the learning ability of our proposed bio-plausible learning rules, we build up a 2-layers SNN to fit a random target spike train (PSC) from randomly generated inputs as shown in Figure \ref{fig:toy_example} (a). 
One SOM cell paired with the output pyramidal cell helps it propagate error backwards and update the input weights. More detail settings can be found in the appendix. In Figure \ref{fig:toy_example} (b) and (c), the upper two sub-figures exploit that both the output $a_i$ and the predict output $a_{i_p}$ are able to fit our randomly assigned sinusoidal target PSC $a_i^{\rm target}$. The lower two sub-figures shows the difference between paired weights, where the two backward weights $\textcolor{NavyBlue}{w_{ji_{pe}}^{(l)}}\approx \textcolor{Red}{w_{ji_e}^{(l)}}$ after training, yet the other pair remains different. Such difference would not hinder the training because the success of error backpropagation only requires $a_i \approx a_{i_p}$. 

\subsection{The Results on the MNIST Dataset and the CIFAR-10 Dataset}
The proposed framework is compared with other BP-based rules on two widely used real-world datasets: MNIST \citep{lecun1998mnist} and CIFAR-10 \citep{krizhevsky2009learning}. Previous works usually use the fixed-step first-order forward Euler method to discretize continuous membrane voltage updates over a set of discrete time steps, we also following this setting and take several measures to guarantee a fair comparison: 1) Mirroring $B(u)$ with $\vartheta$ to get $B(u)$'s value when $u>\vartheta$. 2) Setting $\kappa_{\rm STDP}[t]=1$ when $t=0$, and $\kappa_{\rm STDP}[t]=0$ when $t\neq 0$. 3) All the comparisons are made under a SNN's self-predicting state. The results are concluded in table \ref{tab:MNIST_CIFAR10}. Our method gains comparable performance as compared to BP-based works.

\begin{table}[h]
    \centering
    \vspace{-5mm}
    \caption{Performances comparison of different methods on the MNIST and CIFAR10 datasets}    
    \begin{tabular}{c|c|c|c|c}
        \hline
         & \multicolumn{2}{c|}{MNIST} & \multicolumn{2}{c}{CIFAR10} \\
        \hline
        Method  & \#Steps & BestAcc & \#Steps & BestAcc\\
        \hline
        SLAYER~\citep{shrestha2018slayer}  & 300 & 99.41\%  & null & null\\
        TSSL-BP~\citep{zhang2020temporal}   & 5 & 99.53\%  & 5 & 89.22\%\\
        NA~\citep{yang2021backpropagated}  & 5 & 99.69\% & 5 & 91.76\% \\
        \textbf{This work} & \textbf{5} & \textbf{99.46\%}  & \textbf{5} & \textbf{86.88\%}\\
        \hline
        \multicolumn{5}{l}{MNIST SNN structure: 15C5-P2-40C5-P2-300}\\
        \multicolumn{5}{l}{CIFAR10 SNN structure: 96C3-256C3-P2-384C3-P2-384C3-256C3-1024-1024}
    \end{tabular}
    \label{tab:MNIST_CIFAR10}
    \vspace{-0.5cm}
\end{table}

\section{Conclusion}
We proposed a new bio-plausible learning framework, BioLeaF, consisting of two key components: an architecture, and its paired learning rules. BioLeaF leverages previous bio-plausible works' limitation, and bridges the gap between the bio-plausible approach and the BP-based approach both analytically and experimentally. The equivalence of these two approaches are demonstrated under a special setting, and the comparable experimental performance of them are benchmarked on MNIST and CIFAR10 datasets. This work may provide new insights on both approaches.

\bibliography{iclr2022_conference}
\bibliographystyle{iclr2022_conference}

\newpage
\appendix
\section{Appendix}
\subsection{Full Derivatives of the BP algorithm}

We start from backpropagating the error signal from the PSC - $a$ to the spike train - $s$. A causal impulse response only has value when $t>0$ as the $\epsilon$ in (\ref{eq:Ftype2}). So the partial derivative of a neuron's output spike train on time $t$ is the integration on the derivative of all future error with $t'>t$, which can be expressed by:
\begin{equation}
    \frac{\partial L}{\partial  s_i^{(l)}(t)}= \int_{t'=t}^T \frac{\partial L}{\partial  a_i^{(l)}(t')} \frac{\partial  a_i^{(l)}(t')}{\partial  s_i^{(l)}(t)} dt'
    =\int_{t'=t}^T g_i^{(l)}(t') \epsilon(t'-t) dt' = \left(g_i^{(l)} \ast \tilde{\epsilon}\right)(t),
\label{eq:s_full}
\end{equation}
where $T$ is the length of a simulation time window, and $\tilde{\epsilon}(t) = \epsilon(-t)$ is the time-reverse impulse response kernel of the synapse.

\begin{figure}[b]
    \centering
    \includegraphics[width=0.9\linewidth]{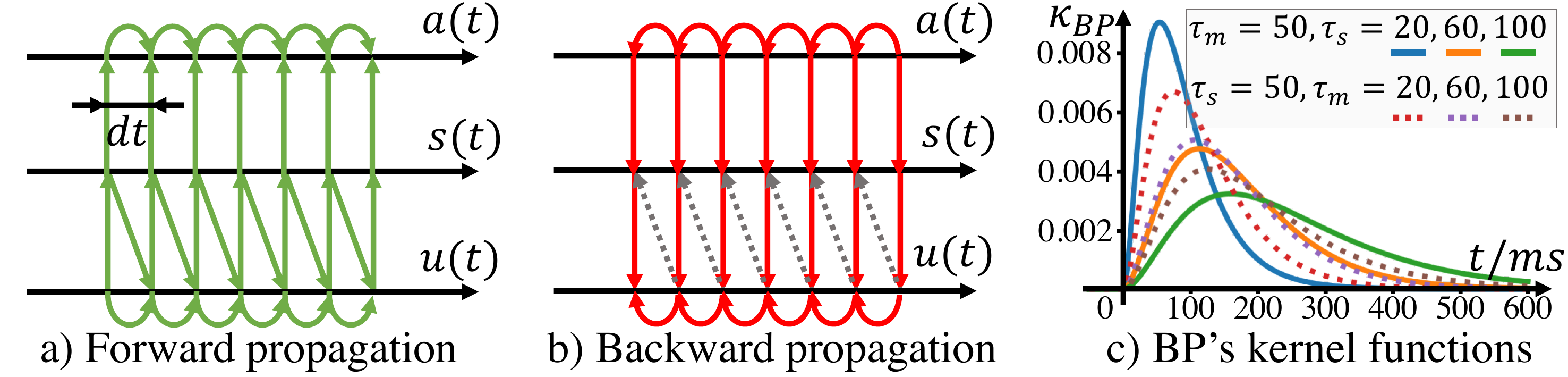}
    \caption{(a) Full forward dependency between $u$, $s$, and $a$ in the discrete simulation. (b) Full backward dependency in the discrete simulation. The dashed lines are usually been omitted when doing backpropagation. (c) BP's kernel functions with different time constants}
    \label{fig:dependency}
\end{figure}

In Figure \ref{fig:dependency}, we draw the full dependency in both (a) - forward and (b) - backward propagation under infinitesimal discrete time steps $dt$. Under which, the gradient on $u$ can be described by:
\begin{equation}
    \frac{\partial L}{\partial u_i^{(l)}(t)} = \frac{\partial L}{\partial s_i^{(l)}(t)} \frac{\partial s_i^{(l)}(t)}{\partial u_i^{(l)}(t)} + \frac{\partial L}{\partial u_i^{(l)}(t+dt)}
    \left(\frac{\partial u_i^{(l)}(t+dt)}{\partial s_i^{(l)}(t)} \frac{\partial s_i^{(l)}(t)}{\partial u_i^{(l)}(t)}+\frac{\partial u_i^{(l)}(t+dt)}{\partial u_i^{(l)}(t)}\right),
\label{eq:u_full}
\end{equation}
which does not have a closed form solution for two reasons:

$\bullet$ The derivative ${\partial s_{i}^{(l)}(t)}/{\partial u_i^{(l)}(t)}$ is zero when $u_i^{(l)}(t) \neq \vartheta$, and is $\infty$ when $u_i^{(l)}(t) = \vartheta$, which is ill defined.

$\bullet$ The complex temporal dependency of  $u_i^{(l)}(t)$ and $u_i^{(l)}(t+dt)$ brings difficulty for calculation.

Following previous works (\# cite): by introducing surrogate gradient as a substitution for ${\partial s_{i}^{(l)}(t)}/{\partial u_i^{(l)}(t)}$, we approx the first term in (\ref{eq:u_full}) as:
\begin{equation}
     \frac{\partial L}{\partial s_i^{(l)}(t)} \frac{\partial s_i^{(l)}(t)}{\partial u_i^{(l)}(t)} \approx \frac{\partial L}{\partial s_i^{(l)}(t)} \sigma'(u_i^{(l)}(t)).
\end{equation}
The other term in (\ref{eq:u_full}) describes the temporal dependency of the $u$. Define a membrane potential's impulse response kernel function $\zeta(t)$ as: $\zeta(t) =\frac{1}{\tau_m}e^{(-t/\tau_m)}H(t)$

By ignoring part of the dependency, as shown by the gray colored dash lines in Figure \ref{fig:dependency} (b), and applying the same trick as in (\ref{eq:s_full}), we have:
\begin{equation}
\begin{aligned}
    \frac{\partial L}{\partial u_i^{(l)}(t)} \approx \int_{t'=t}^{T}\frac{\partial L}{\partial s_i^{(l)}(t')} &\frac{\partial s_i^{(l)}(t')}{\partial u_i^{(l)}(t')}\frac{\partial u_i^{(l)}(t')}{\partial u_i^{(l)}(t)} dt\approx \int_{t'=t}^{T} \frac{\partial L}{\partial s_i^{(l)}(t')}  \sigma'(u_i^{(l)})(t') \zeta(t'-t)\\
    &= \left[\left((g_i^{(l)} \ast \tilde{\epsilon}) \cdot \sigma'(u_i^{(l)})\right) \ast \tilde{\zeta}\right] (t),
\label{eq:loss_to_u}
\end{aligned}
\end{equation}
where $\tilde{\zeta}(t) = \zeta(-t)$ is the time-reverse impulse response kernel of the membrane potential.

Then the final step is to propagate the gradient from $u_i^{(l)}$ to $\textcolor{ForestGreen}{w_{ij}^{(l)}}$ according to (\ref{eq:pyramidal}):
\begin{equation}
\begin{aligned}
    \frac{\partial L}{\partial \textcolor{ForestGreen}{w_{ij}^{(l)}}} &= \int\frac{\partial L}{\partial u_i^{(l)}(t)}\frac{\partial u_i^{(l)}(t)}{\partial \textcolor{ForestGreen}{w_{ij}^{(l)}}} dt 
    \approx \int \left[\left((g_i^{(l)} \ast \tilde{\epsilon}) \cdot \sigma'(u_i^{(l)})\right) \ast \tilde{\zeta}\right](t)\cdot a_j^{(l-1)}(t) dt\\
    &\approx \int \left((g_i^{(l)}\cdot \sigma'(u_i^{(l)})) \ast \tilde{\epsilon} \ast \tilde{\zeta} 
    \right)(t)\cdot a_j^{(l-1)}(t) dt \\
    &= \int \left(g_i^{(l)}\cdot \sigma'(u_i^{(l)})\right)(t) \cdot\left(a_j^{(l-1)} \ast \epsilon \ast \zeta \right)(t) dt \\
    &= \int e_i^{(l)}(t) \cdot\left(s_j^{(l-1)}\ast \kappa_{\rm BP} \right)(t)  dt,
\end{aligned}
\label{eq:BP_rule}
\end{equation}
where $e_i^{(l)}(t)=(g_i^{(l)}\cdot \sigma'(u_i^{(l)}))(t)$, and $\kappa_{\rm BP}(t)=(\epsilon\ast\epsilon\ast \zeta)(t)$. We show the shape of $\kappa_{\rm BP}$in Figure \ref{fig:dependency} (c), and calculate its close form expression as:
\begin{equation}
    \kappa_{\rm BP}(t)=
    \left[\frac{\tau_m \cdot (e^{-\frac{t}{\tau_m}}-e^{-\frac{t}{\tau_s}})}{(\tau_m-\tau_s)^2} -\frac{t \cdot e^{-\frac{t}{\tau_s}}}{\tau_s(\tau_m-\tau_s)}\right]\cdot H(t).
\end{equation}
We then provide the rules to further propagate the gradient to previous layers (take layer $(l-1)$ as an example). Following (\ref{eq:pyramidal}), we have:
\begin{equation}
    g_j^{(l-1)} =\frac{\partial L}{\partial  a_j^{(l-1)}(t)}=\sum_{i=1}^{N^{(l)}} \textcolor{ForestGreen}{w_{ij}^{(l)}}\frac{\partial L}{\partial u_i^{(l)}(t)}=\sum_{i=1}^{N^{(l)}} \textcolor{ForestGreen}{w_{ij}^{(l)}} \left[\left((g_i^{(l)} \ast \tilde{\epsilon}) \cdot \sigma'(u_i^{(l)})\right) \ast \tilde{\zeta}\right] (t)
\end{equation}
BP in hidden layers also following the similar steps: from $g_j^{(l-1)}$ to $\partial L/\partial u_j^{(l-1)}$ following (\ref{eq:loss_to_u}), and further propagating to weights following (\ref{eq:BP_rule}).

\subsection{Detailed Experimental Setups}
\subsubsection{Universal Spike Train Approximator}
This section concludes all settings used in the 2-layers SNN experiment. All parameters used are summarized in Table \ref{tab:2-layers}.

\begin{table}[h]
    \centering
    \vspace{-0.5cm}
    \caption{Parameters of the 2-layers network.}    
    \begin{tabular}{c c c c c c c }
         \hline
         $N_T$ &$\tau_m$ &$\tau_s$ & \#inputs & \#hidden & $p_{in}$ & $A^+$ \\ 
         500ms & 50ms & 20ms & 50 & 100 & 0.05 & 0.00004 \\
         \hline
         $\tau_+$ & $A^-$ & $v_{\rm rest}$ & $\vartheta$ &\#iters  &  scale & bias\\
         30ms & 0& 0 &1 & 5000  &  0.3 & 0.01\\
         \hline
    \end{tabular}
    \label{tab:2-layers}
\end{table}

\begin{figure}[b]
    \centering
    \includegraphics[width=0.999\linewidth]{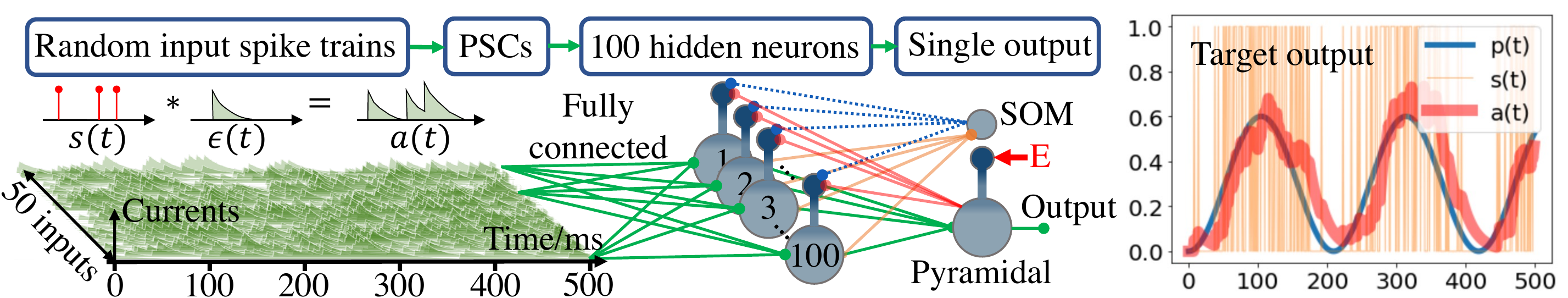}
    \caption{Visualization of all 50 input currents, the network architecture, and the process to generate the target output current}
    \label{fig:toy_inputs}
\end{figure}

The total simulation time $N_t$ is 500ms. $\tau_m$ and $\tau_s$ are the time constants in (\ref{eq:pyramidal}) and (\ref{eq:g}). The total number of randomly generated inputs is 50. The number of hidden pyramidal cells is 100. 

We sample the total 500ms by 1ms when doing this experiment, which means there are 500 time steps totally. As shown in Figure \ref{fig:toy_inputs}, the randomly generated input currents are produced by a two-steps process.

The first step is to generate spike trains $s(t)$ following Bernoulli distribution on each time step with probability $p_{in}=0.05$. Then the input currents are defined using F-type synapses $a(t) = s(t) \ast \epsilon$ as in (\ref{eq:Ftype2}). 

$A^+$, $\tau_+$, and $A^-$ are parameters of our STDP function. $v_{\rm rest}$ and  $\vartheta$ are fixed to 0 and 1 in all experiment of our work. All weights are initialized by a scaled Gaussian distribution $w\sim\left({\rm scale}\times N(0,1) + {\rm bias}\right)$. 

The output target signal is defined by a sinusoidal probability function $p_{\rm target}(t) = 0.3 + -0.3cos(0.03t)$, and $p_{\rm target}(t)$ is converted to spike trains $s_{\rm target}(t)$ following the Bernoulli distribution on each time step, further, $s_{\rm target}(t)$ is converted to continuous currents following $a(t) = s(t) \ast \epsilon$. The reason we do such conversion is to guarantee that the single output neuron is possible to fit the target output signal perfectly.

\subsubsection{MNIST}
The experiment runs on a single RTX-3090 GPU. The MNIST dataset \citep{lecun1998mnist} has 60,000 training images and 10,000 testing images. The training parameters are: batch size = 64, number of training epochs = 200, and learning rate = 0.0005 for the adopted AdamW optimizer \citep{loshchilov2017decoupled}. The images were converted to  continuous-valued multi-channel  currents.    Moreover, data augmentations including RandomCrop and RandomRotation were applied to improve performance \citep{shorten2019survey}. 

\subsubsection{CIFAR10}
The experiment runs on a single RTX-3090 GPU. The CIFAR10 dataset \citep{krizhevsky2009learning} has 50,000 training images and 10,000 test images. We trained our SNN using BioLeaF for 1200 epochs with a batch size of 50 and a learning rate 0.0005 for the AdamW optimizer \citep{loshchilov2017decoupled}. 
The same input image coding strategy as the MNIST dataset was adopted. Moreover, data augmentations including RandomCrop, ColorJitter, and RandomHorizontalFlip \citep{shorten2019survey} were applied. The convolutional layers were initialized using the kaiming uniform initializer \citep{he2015delving}, and the linear layers were initialized using the kaiming normal initializer \citep{he2015delving}.

\subsection{Architecture illustration}
\begin{figure}[t]
    \centering
    \includegraphics[width=0.95\linewidth]{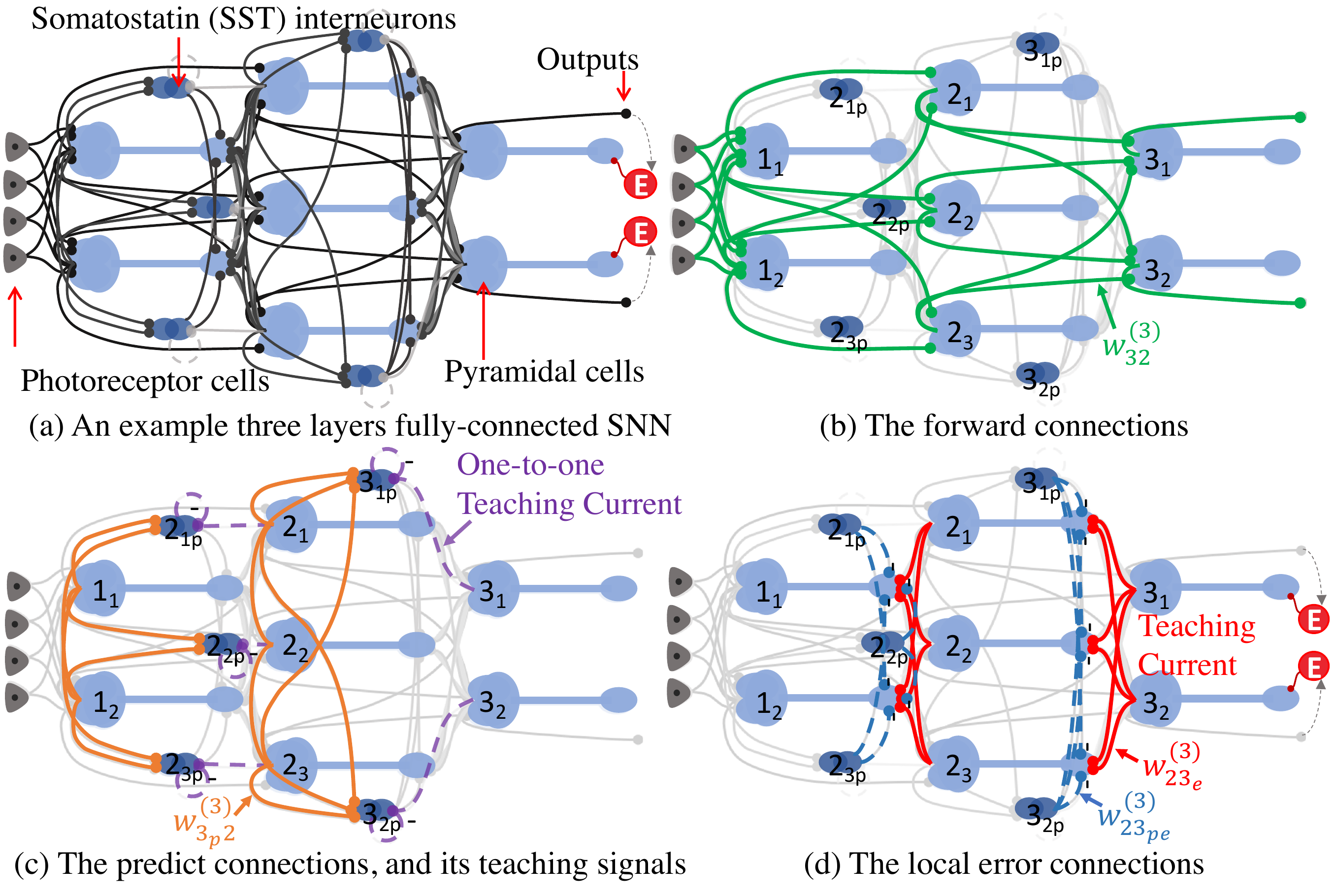}
    \caption{Our proposed Microcircuit architecture.}
    \label{fig:framework2}
\end{figure}

We provide more figures to illustrate our proposed microcircuit architectures. Figure \ref{fig:framework2} (a) is another example of our spiking neural networks' architecture: Two types of cells, Pyramidal cells and Somatostatin (SOM) cells, are needed to realize the forward propagation and the local synaptic plasticity. This example three layers fully-connected (FC) neural network has 2-3-2 pyramidal cells in each layer. The input current signals are from four photoreceptor cells in this example simulating a vision related learning task. 

(b) to (d) are the disassembled explanation of all synaptic connections. Each neuron has been indexed by its layer and the footnoted position in its layer. The pyramidal cells and SOM cells are one-on-one in each layer except for the $1^{\rm st}$ layer, where no SOM cells are needed.

$\bullet$  (b) The connections in this sub-Figure (green colored solid arrows) are all feed-forward connections as the same as the weight connections in the more conventional non-spiking artificial neural networks (ANNs). 

$\bullet$  (c) The output currents of the pyramidal cells in a layer are connected to the SOM cells in the next layer (orange colored solid arrows). The SOM cells use these signals to predict the firing activity of pyramidal cells in the next layer, so we name these connections as predict connections, and use the footnote $p$ to represent all of the predict-related parameters. The adjustment of these predict connections requires all SOM cells to receive the one-to-one teaching current signal from each SOM cell's corresponding pyramidal cell (purple colored dashed arrows).

$\bullet$  (d) The top-down feed backward signals (red colored solid arrows), carrying the superposition of output current from the next layer and error information, are connected to the apical dendrites of previous layer's pyramidal cells. The next layer's output current signals should be canceled out by the local predicting signals that the next layer's SOM cells provided (blue colored dashed arrows), which will leave only error signals on the pyramidal cells' apical dendrites. Since these two types of connections together generate pyramidal cells' \emph{error} information, we use the footnote $e$ to mark all the error related variables. Importantly, the output layer's pyramidal cells needs additional error signals $e_i=a_i^{\rm target}-a_i$ connected to its apical dendrites. Such error signals may come from higher to lower brain areas \citep{leinweber2017sensorimotor}.

\begin{figure}[h]
    \centering
    \includegraphics[width=0.95\linewidth]{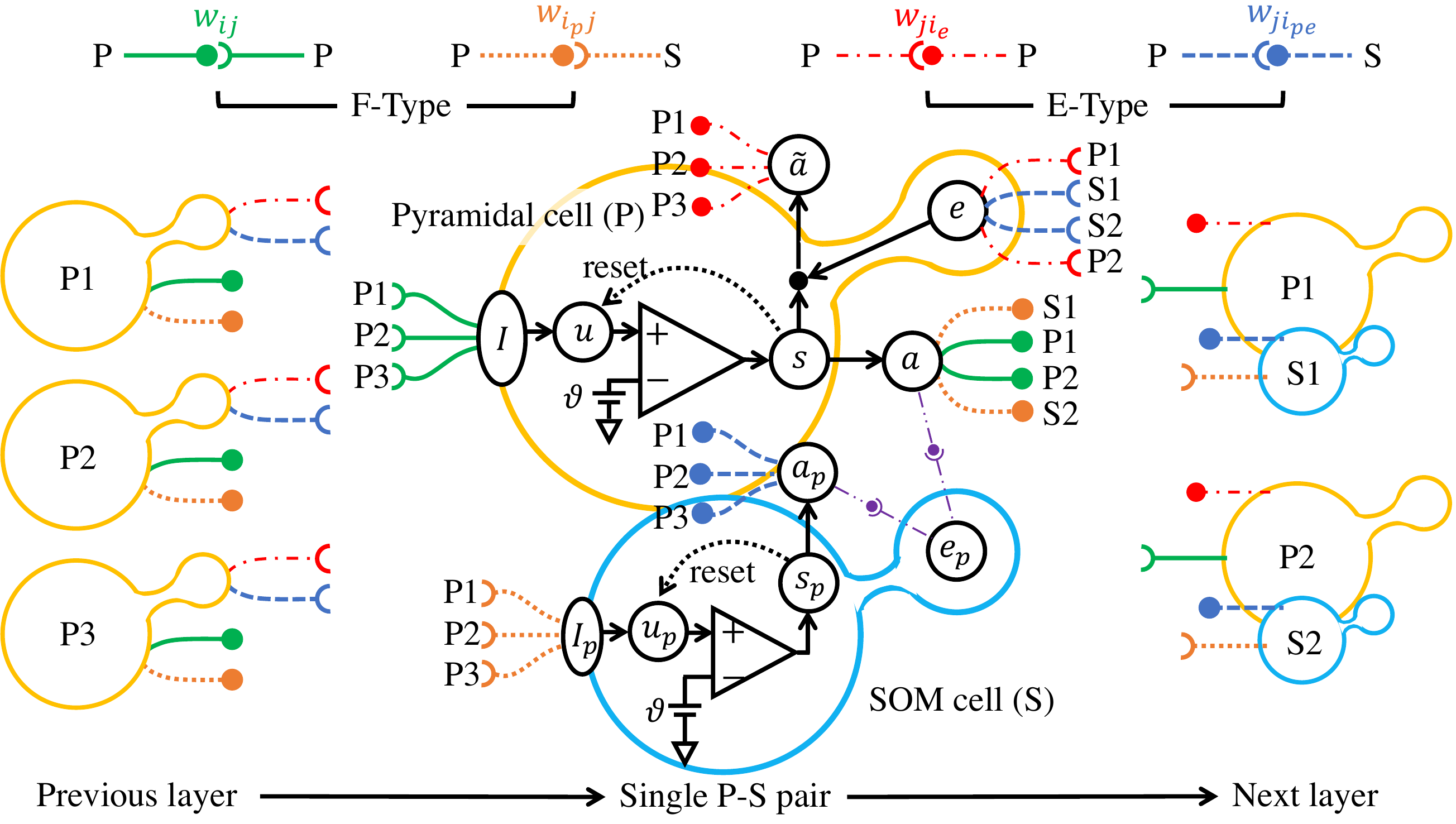}
    \caption{Our proposed Microcircuit architecture.}
    \label{fig:framework3}
\end{figure}

Figure \ref{fig:framework3} zoomed into one pair of Pyramidal-SOM cells, which has three input pyramidal cells from its previous layer, and forward connected to two pairs of Pyramidal-SOM cells in its next layer. Cells in a same layer are indexed, which is used to indicate synapses connections.

\end{document}